\pdfoutput=1
\documentclass[10pt,twocolumn,letterpaper]{article}
\usepackage{iccv}
\usepackage{times}
\usepackage{epsfig}
\usepackage{graphicx}
\usepackage{amsmath}
\usepackage{amssymb}
\usepackage[hmargin=2cm,vmargin=2.5cm]{geometry}
\usepackage{etoolbox}
\usepackage{multirow}
\usepackage{longtable}%
\usepackage{colortbl}%
  
\usepackage{booktabs}

\newif\ifblackandwhite
\usepackage[table]{xcolor}
\definecolor{light-gray}{gray}{0.95}

\usepackage{tikz}
\definecolor{gold}{rgb}{1.0, 0.84, 0.0}
\definecolor{silver}{rgb}{0.75, 0.75, 0.75}
\definecolor{bronze}{rgb}{0.8, 0.5, 0.2}

\newcommand{\gold}[0]{\tikz\draw[gold,fill=gold] (0,0) circle (.5ex);}
\newcommand{\silver}[0]{\tikz\draw[silver,fill=silver] (0,0) circle (.5ex);}
\newcommand{\bronze}[0]{\tikz\draw[bronze,fill=bronze] (0,0) circle (.5ex);}

\ifblackandwhite
  
\else

\fi

\usepackage[breaklinks=true,bookmarks=false]{hyperref}

\iccvfinalcopy 

\def\iccvPaperID{****} 

\ificcvfinal\fi

\begin{document}

\title{Instant Continual Learning of Neural Radiance Fields}

\author{Ryan Po \;\;\;\;\;\;\;\;\; Zhengyang Dong \;\;\;\;\;\;\;\;\; Alexander W. Bergman \;\;\;\;\;\;\;\;\; Gordon Wetzstein
\\
Stanford University\\
{\tt\small {\{rlpo, awb, leozdong, gordonwz\}}@stanford.edu}
}

\maketitle
\ificcvfinal\thispagestyle{empty}\fi

\begin{abstract}
   Neural radiance fields (NeRFs) have emerged as an effective method for novel-view synthesis and 3D scene reconstruction. However, conventional training methods require access to all training views during scene optimization. 
   This assumption may be prohibitive in continual learning scenarios, where new data is acquired in a sequential manner and a continuous update of the NeRF is desired, as in automotive or remote sensing applications.
   When naively trained in such a continual setting, traditional scene representation frameworks suffer from catastrophic forgetting, where previously learned knowledge is corrupted after training on new data. Prior works in alleviating forgetting with NeRFs suffer from low reconstruction quality and high latency, making them impractical for real-world application. We propose a continual learning framework for training NeRFs that leverages replay-based methods combined with a hybrid explicit--implicit scene representation. Our method outperforms previous methods in reconstruction quality when trained in a continual setting, while having the additional benefit of being an order of magnitude faster. 
\end{abstract}

\section{Introduction}
High-quality reconstruction and image-based rendering of 3D scenes is a long-standing research problem spanning the fields of computer vision \cite{Janai2017ComputerVF,Ma2019AccurateM3}, computer graphics \cite{Bruno2010From3R,Gafni2020DynamicNR}, and robotics \cite{Bailey2006SimultaneousLA, DurrantWhyte2006SimultaneousLA, MurArtal2015ORBSLAMAV}. Recently, the introduction of Neural Radiance Fields (NeRFs) \cite{Mildenhall2021NeRF} has led to substantial improvements in this area through the use of differentiable 3D scene representations supervised with posed 2D images. 
However, NeRFs require access to all available views of the 3D scene during training, a condition that is prohibitive for automotive and remote sensing applications, among others, where data is sequentially acquired and an updated 3D scene representation should be immediately available. 
In such conditions, the scene representation must be trained in a continual setting, where the model is given access to a limited number of views at each stage of training, while still tasked with reconstructing the entire scene.
    
\begin{figure}[t]
    \centering
    \includegraphics[width=\linewidth]{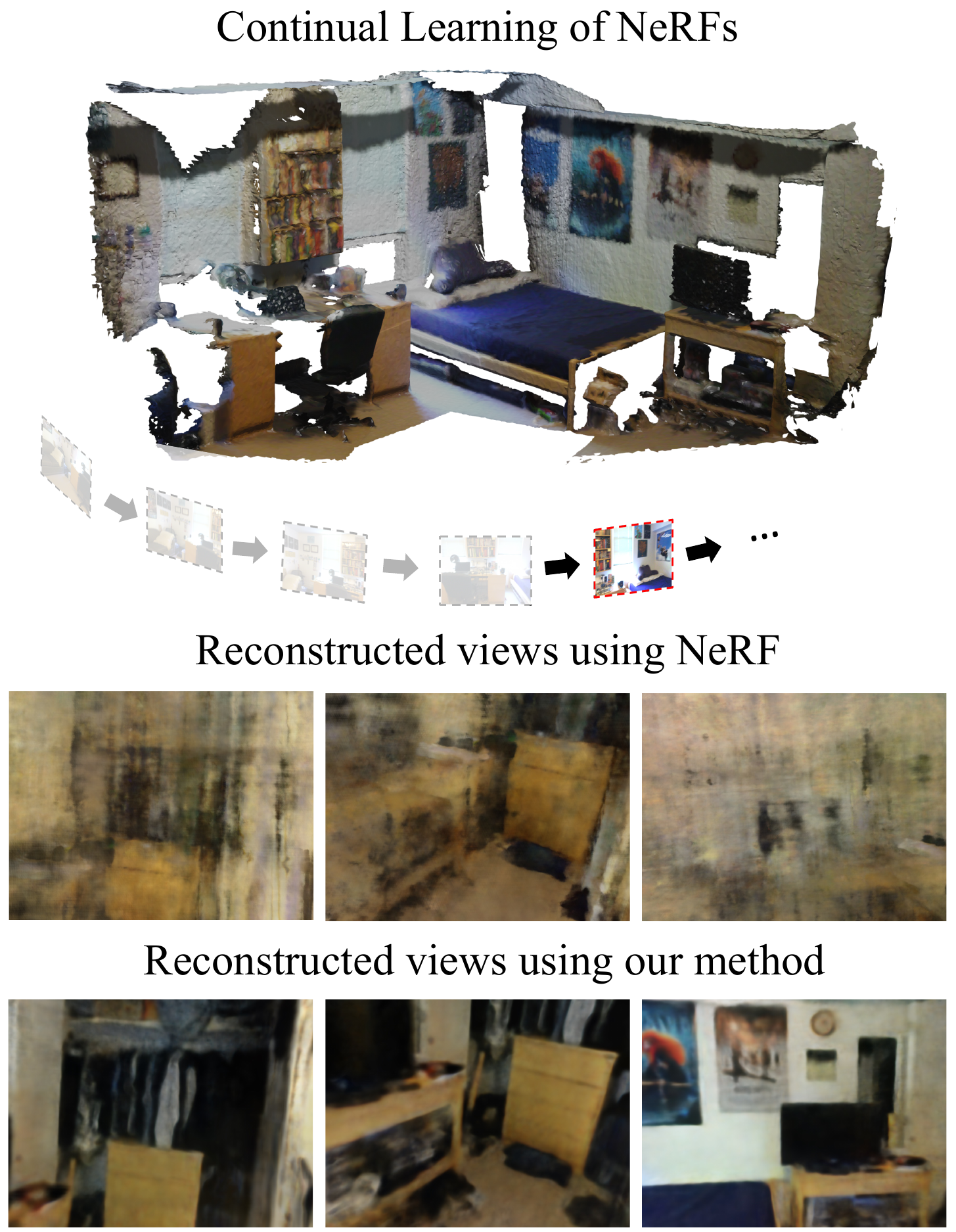}
    \caption{\textbf{Continual learning of NeRFs.} Conventionally, NeRFs are trained with access to all training views. However, for continual learning scenarios we must train on batches of input views without access to previously seen data (top). When trained in these settings, conventional methods suffer from catastrophic forgetting, leading to poor reconstructions (center). In contrast, our method reconstructs the entire scene with high quality (bottom).}
    \label{fig:teaser}
    \vspace{-1.5em}
\end{figure}

\begin{figure}[t]
    \centering
    \includegraphics[width=\linewidth]{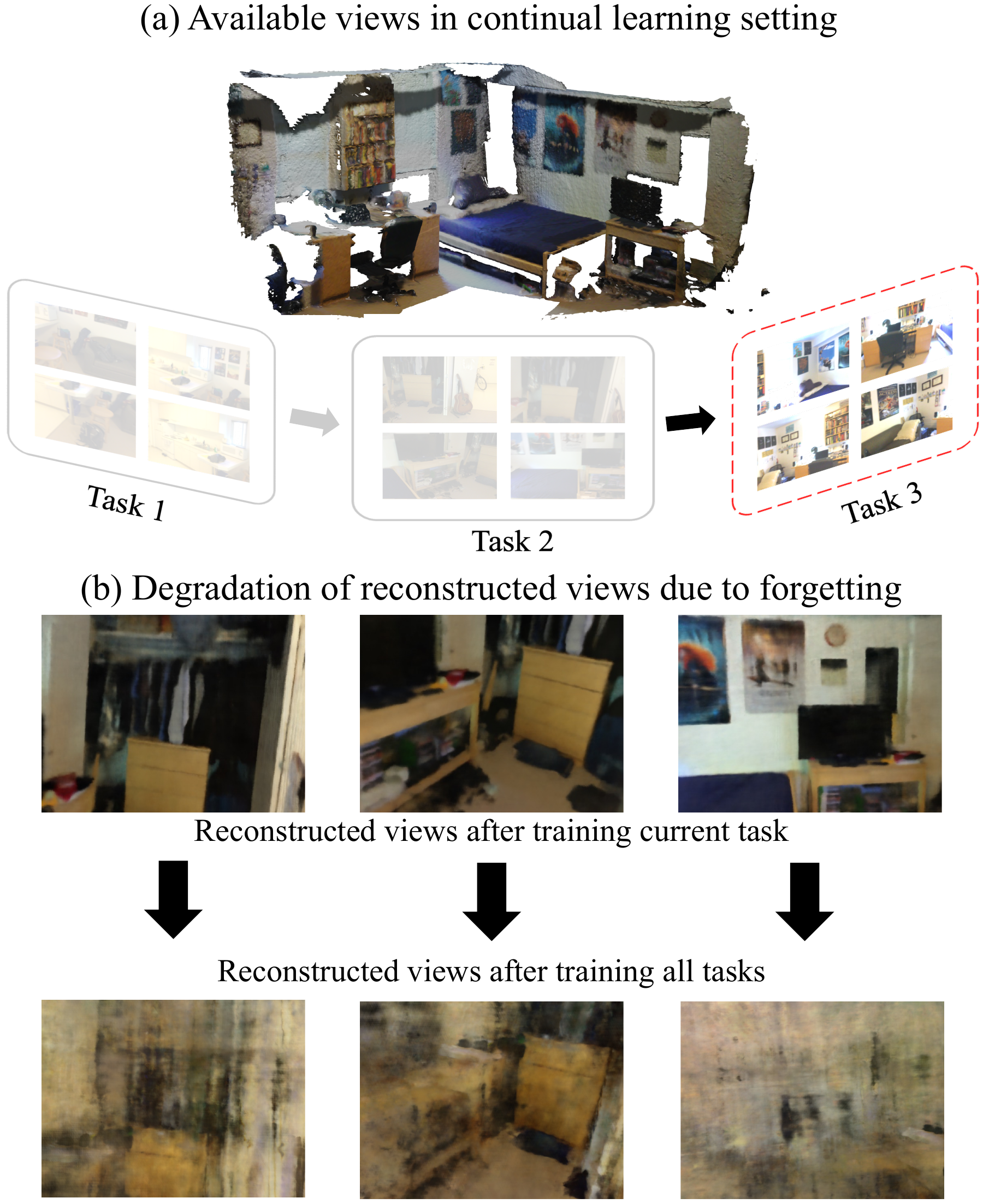}
    \caption{\textbf{Problem overview.} (a) Continual learning setting for training NeRFs. Instead of training the scene representation over all input views at once, the model is given 2D views of the scene in sequential batches. During a particular stage in training, the model is only given access to the most recently captured views. (b) Training NeRF is the continual setting leads to catastrophic forgetting. Previously learned 3D scene content is corrupted after training on newly captured views.}
    \label{fig:overview}
\end{figure}

When trained in a continual setting, NeRFs suffer from catastrophic forgetting \cite{French1999CatastrophicFI}, where previously learned knowledge is forgotten when trained on new incoming data. Recent work \cite{Sucar2021iMAPIM, Chung2022MEILNeRFMI} has shown promise in tackling catastrophic forgetting through replay-based methods. Such approaches aim to alleviate forgetting by storing information from previous tasks either explicitly or in a compressed representation, then revisiting this information during training of subsequent tasks. Existing methods \cite{Zhu2021NICESLAMNI,Rosinol2022NeRFSLAMRD} have seen success through the application of replay-based techniques in conjunction with NeRF for addressing the task of simultaneous mapping and localization (SLAM) \cite{Bailey2006SimultaneousLA}, however such methods either suffer from memory scalability or latency issues.

In this work, we tackle the task of continually learning NeRFs by leveraging the benefits of replay-based techniques. Specifically, we acknowledge that a trained NeRF itself is a compressed representation of all previously observed 2D views. By freezing a copy of the scene representation after the training of each task, we essentially have access to pseudo ground truth RGB values for all previously seen data by querying this oracle. We also modify the underlying neural scene representation architecture motivated by one key insight: catastrophic forgetting is a fundamental problem faced by neural networks. Therefore, the fully implicit (MLP) representation used by NeRF is fundamentally ill-suited for the task of continual learning. We minimize the reliance of our underlying scene model on the decoder neural network by using a hybrid implicit--explicit representation. By replacing the frequency encoding in NeRF with a multi-resolution hash encoding \cite{Mller2022InstantNG}, we greatly reduce the size of the decoder multilayer perceptron (MLP), minimizing the effects of catastrophic forgetting.

As an additional benefit, our method is also an order of magnitude faster than previous replay-based methods \cite{Chung2022MEILNeRFMI}. This enables fast continual scene fitting, as our method can learn additional 3D scene content from new input views in as little as 5 seconds (see Section \ref{results} for details). 
\section{Related Work}

\paragraph{Neural radiance fields.}

Scene representation networks~\cite{sitzmann2019scene} and neural rendering~\cite{tewari2020state,tewari2022advances} have emerged as a family of techniques enabling effective 3D scene reconstruction. 
Given a set of images and corresponding ground truth camera poses, neural radiance fields (NeRFs)~\cite{Mildenhall2021NeRF}, for example, optimizes a underlying scene representation by casting rays, 
sampling the scene volume and aggregating sampled color and density values to synthesize an image. The success of NeRFs has spawned a line 
of works on improving the quality and efficiency of the method~\cite{Barron2021MipNeRF3U,Barron2021MipNeRFAM,Chen2021MVSNeRFFG,Hedman2021BakingNR,Kellnhofer2021NeuralLR,bergman2021metanlr,Lin2021BARFBN,Liu2020NeuralSV,MartinBrualla2020NeRFIT,Mller2022InstantNG,Reiser2021KiloNeRFSU,Tancik2020LearnedIF,Tancik2020FourierFL,Verbin2021RefNeRFSV,Wang2021IBRNetLM,Yang2021BANMoBA,Yu2021PlenoxelsRF,Yu2020pixelNeRFNR}, 
while extending the method to a range of applications~\cite{Xie2021NeuralFI,Chen2021AnimatableNR,Lombardi2021MixtureOV,Peng2021AnimatableNR,Hao2021GANcraftU3,Ichnowski2021DexNeRFUA,Oechsle2021UNISURFUN,Sucar2021iMAPIM,Zhu2021NICESLAMNI}.
NeRFs leverage a neural implicit representation (NIR)~\cite{Sitzmann2020ImplicitNR} in the form of a simple, yet effective multi-layer pereceptron (MLP) 
to represent the 3D scene. Many follow-up works improve on the underlying NIR, enabling features such as real-time rendering~\cite{Reiser2021KiloNeRFSU,Yu2021PlenOctreesFR,Chan2021EfficientG3} and
faster training~\cite{Mller2022InstantNG,Liu2020NeuralSV,Yu2021PlenoxelsRF,Chen2022TensoRFTR}. A key limitation for the training of NeRFs is the assumption
that all input images of the target scene are available during training. In scenarios such as autonomous vehicle or drone footage captures, 
this assumption no longer holds as data is sequentially acquired and an updated 3D representation should be immediately available. NeRFs trained
on sequential data suffer from catastrophic forgetting~\cite{Robins1995CatastrophicFR}. Our method overcomes this limitation, providing a high
quality reconstruction of the entire scene, while imparting minimal computational and memory overhead.

\begin{figure*}[t]
    \centering
    \includegraphics[width=\linewidth]{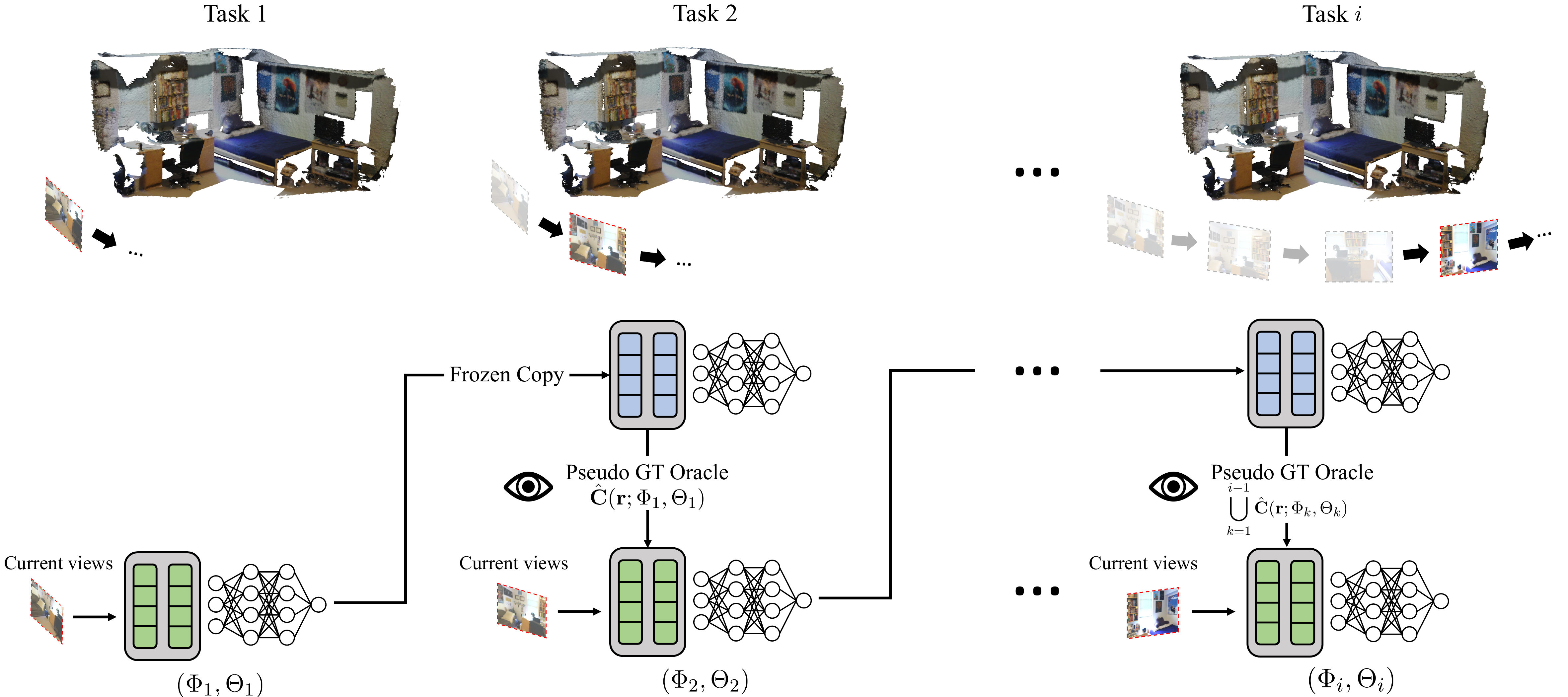}
    \caption{\textbf{Memory replay through NeRF distillation.} Scene representation is sequentially trained on sequentially acquired views. After each stage of training, a frozen copy of the scene parameters is stored. While optimizing for the next set of incoming images, the frozen network is queried to obtain pseudo ground truth values. The current network $(\Phi_i, \Theta_i)$ is trained on a mixed objective that minimizes photometric loss with respect to ground truth images from the current task, and pseudo ground truth values for previous tasks (Equation~\ref{loss}).}\label{fig:method}
\end{figure*}

\paragraph{Continual learning.} Continual learning is a long-standing problem in the field of machine learning, 
where partial training data is available at each stage of training. As mentioned above, NeRFs trained in a continual
learning setting suffers from catastrophic forgetting~\cite{Robins1995CatastrophicFR}. Existing work in this field fall into
three main categories~\cite{DeLange2019ACL}: parameter regularization~\cite{Li2016LearningWF,Triki2017EncoderBL,Aljundi2017MemoryAS,Kirkpatrick2016OvercomingCF}, 
parameter isolation~\cite{Aljundi2016ExpertGL,Xu2018ReinforcedCL,Mallya2017PackNetAM,Fernando2017PathNetEC} and 
data replay~\cite{Klein2007ParallelTA,Rebuffi2016iCaRLIC,Rolnick2018ExperienceRF,Shin2017ContinualLW,LopezPaz2017GradientEM,Chaudhry2018EfficientLL}.
Parameter isolation methods aim at combating catastrophic forgetting by attempting to learn a sub-network for each task, while parameter regularization methods identify parameters important
for preserving old knowledge and penalizing changes to them. Finally, data replay methods preserve previous knowledge by storing a subset of
previous training data. Subsequent tasks are then optimized over old and new incoming data. Our proposed method leverages a self-distillation method
similar to previous data replay approaches, storing pseudo ground truth values for all previous training with minimal memory usage.

\paragraph{SLAM \& continual learning of NeRFs.} Works in the field of simultaneous mapping and localization (SLAM)~\cite{Bailey2006SimultaneousLA}
aim at reconstructing a 3D scene from a continuous stream of images, similar to the continual learning setting. Recent works 
~\cite{Sucar2021iMAPIM,Zhu2021NICESLAMNI,Rosinol2022NeRFSLAMRD} combine NIRs and traditional SLAM-based methods with promising results.
These methods fall under the data replay category, as they approach the task of continual learning by explicitly storing key-frames from 
previous image streams. Storing data explicitly can be expensive, and designing an appropriate importance heuristic for selecting key-frames
is non-trivial. In contrast, our approach stores previous data as an implicitly defined generator, greatly reducing memory overhead.

\section{Continual Learning of NeRFs}
Before we explain the details of our proposed method, it is important to first formally establish the task of continual learning of NeRFs. We consider the scenario where $t$ sets of image data come in sequentially, represented by $\{\mathcal{I}_1,\dots,\mathcal{I}_t\}$. Each image data set is represented by $\mathcal{I}_i = (\mathbf{I}_i, \mathbf{R}_i)$, where $\mathbf{I}_i$ represents the per-pixel RGB values of the image data and $\mathbf{R}_i$ represent the camera rays corresponding to each image pixel. Note that $\mathbf{R}_i$ can either be explicitly stored as values in $\mathbb{R}^6$ (ray origin and direction) or implicitly through camera extrinsic and intrinsic matrices. 

The objective of our optimization remains the same, we wish to minimize reconstruction loss across all provided ground truth views in $\{\mathcal{I}_1,\dots,\mathcal{I}_t\}$. However, the training procedure differs from conventional NeRF training. Training is performed sequentially as illustrated in Figure \ref{fig:overview}a. At a given stage of training, our model is only given access to a subset of all of the RGB images (visualized in Figure \ref{fig:overview}b), but access to ray information from all previous tasks. Formally, at time step $i$, the model is able to access $\mathbf{I}_i$ and $\{\mathbf{R}_1,\dots,\mathbf{R}_{i}\}$. Note that this formulation is slightly different from prior works such as MEIL-NeRF~\cite{Chung2022MEILNeRFMI} where access to ray information is also constrained. However, we believe this constraint is unwarranted since ray information can be stored implicitly for every input view with only 6 scalar values\footnote{Camera extrinsic matrices can be implicitly stored in the form $(t_x, t_y, t_z, r_x, r_y, r_z)$, where $t_x, t_y, t_z$ represents the position of the camera optical center and $r_x, r_y, r_z$ the orientation of the camera.}, assuming all input views share the same camera intrinsics. Similar to prior work~\cite{Chung2022MEILNeRFMI}, our method is based on self distillation~\cite{Hinton2015DistillingTK}, therefore we also assume that we have access to a frozen copy of our trained representation from the previous task.

\section{Method}
In this section, we first provide a brief recap behind the formulation of NeRFs~\cite{Mildenhall2021NeRF}, then introduce our solution to catastrophic forgetting in the context of training NeRFs in a continual setting. There are two main contributors to our solution: namely, the use of a hybrid feature representation (Section~\ref{ngp}) and task specific network distillation (Section~\ref{distillation}).

\subsection{NeRF preliminaries}\label{nerf}
Neural radiance fields (NeRFs)~\cite{Mildenhall2021NeRF} represent a 3D scene through an implicit function from a point in 3D space $\mathbf{x} = (x,y,z)$ along with a corresponding viewing direction $\mathbf{d} = (\theta, \phi)$ to a density value $\mathbf{\sigma}$ and RGB color $\mathbf{c} = (r,g,b)$. Conventionally, NeRFs are represented with an MLP characterized by its parameters $\Theta$, giving the mapping
\begin{equation}
    F_\Theta: (\mathbf{x}, \mathbf{d}) \mapsto (\mathbf{\sigma}, \mathbf{c}).
\end{equation}
Novel views of the 3D scene are generated through volume rendering~\cite{Kutulakos1999ATO} of the 5D radiance field. Given an image pixel with the corresponding ray $\mathbf{r} = (\mathbf{r}_o, \mathbf{r_d})$, by sampling points $\mathbf{x}_i$ along this ray and evaluating the radiance field values $(\mathbf{\sigma}_i, \mathbf{c}_i)$ at these points, the color associated with this ray can be recovered. With $\mathbf{N}$ sampled points, the RGB color of a ray $\mathbf{r}$ is obtained by
\begin{equation}
    \hat{\mathbf{C}}(\mathbf{r};\Theta) = \sum_{i=1}^N T_i(1 - \text{exp}(-\sigma_i\delta_i))\mathbf{c}_i,
    \label{eq:vol_render}
\end{equation}
where $\delta_i$ represents the distance between the $i^{th}$ and $(i+1)^{th}$ sampled point and $T_i$ represents the accumulated transmittance from $r_o$ to the current sample point, given by $T_i = \text{exp}\left(-\sum_{j=1}^{i-1} \sigma_j \delta_j\right)$.

\subsection{Multi-resolution hash encoding}\label{ngp}
Prior work~\cite{Mller2022InstantNG} has found success in replacing the fully implicit $F_\Theta$ with a hybrid representation, leading to faster convergence rates along with better memory and computational efficiency. Hybrid representations map 3D coordinates to an explicitly defined feature space before passing these features into a significantly smaller implicit MLP decoder to obtain density and RGB values. We leverage these explicit feature mappings to alleviate the effects of catastrophic forgetting.
\rowcolors{2}{light-gray}{}
\begin{table*}[t]
\caption{\textbf{Quantitative results: unconstrained setting.} PSNR of different continual learning methods. Every method is trained on each task until convergence, which differs by method. Approximate training time for all 10 tasks is listed next to each method. For each scene, we mark the best performing methods with gold {\protect\tikz\protect\draw[gold,fill=gold] (0,0) circle (.5ex);}, silver {\protect\tikz\protect\draw[silver,fill=silver] (0,0) circle (.5ex);} and bronze {\protect\tikz\protect\draw[bronze,fill=bronze] (0,0) circle (.5ex);} medals. Results marked with * are trained in a non-continual setting, where ground truth data from all tasks are available during scene optimization. These results serve as an upper bound for scenes trained in a continual setting.  Our method consistently out-performs all baselines while taking significantly less time to converge.}
  \centering
  \begin{tabular}{m{10em}m{4em}m{3.5em}m{3.5em}m{3.5em}m{4.5em}m{3.5em}m{3.5em}m{3.5em}}
  \toprule
    & \multicolumn{3}{c}{\textit{ScanNet}} & \multicolumn{3}{c}{\textit{Tanks \& Temples}} & \multicolumn{2}{c}{\textit{TUM RGB-D }}\\
     \;\;\textit{Method} & \textit{0101}  & \textit{0146} & \textit{0160} & \textit{Truck} & \textit{Caterpillar} & \textit{Family} & \textit{Desk 0} & \textit{Desk 1}\\
    \midrule
    \;\;NeRF-Incre (2 hours) & 13.70 & 13.20 & 17.31  & 16.88 \bronze& 15.36 \bronze& 22.96 \bronze& 13.05 \bronze& 14.03 \\
    \;\;iNGP-Incre (10 min) & 16.51 \bronze& 16.64 & 19.98 & 13.49 & 14.55 & 21.15 & 12.70 & 14.65 \bronze\\
    \;\;iNGP + EWC (10 min) & 16.11 & 17.32 \bronze& 20.16 \bronze& 12.50 & 13.61 & 19.28 & 12.50 & 10.85 \\
    \;\;MEIL-NeRF (2 hours) & 24.32 \silver& 26.82 \silver& 28.93 \silver& 22.74 \gold& 20.89 \silver& 26.57 \silver& 20.79 \gold& 19.80 \silver\\
    \;\;Ours (10 min) & 25.72 \gold& 27.87 \gold& 30.28 \gold& 22.71 \silver& 22.51 \gold& 29.33 \gold& 20.65 \silver& 20.34 \gold\\
    \midrule
    \;\;NeRF* (2 hours) & 26.15 & 28.48 & 30.88  & 24.80 & 23.14 & 29.33 & 22.35 & 20.88\\
    \;\;iNGP* (10 min) & 26.00 & 28.43 & 31.16 & 24.22 & 24.02 & 31.14 & 20.95 & 20.73 \\
    \bottomrule
  \end{tabular}\label{tab:main_results}
\end{table*}
\paragraph{Multi-resolution feature grids.}Following Instant-NGP~\cite{Mller2022InstantNG}, we map 3D coordinates to explicit features arranged into $L$ levels, each level containing a maximum of $T$ features, with each feature having a dimensionality of $F$. Each level stores features corresponding to vertices of a 3D grid with fixed resolution. Consider a single feature level $l$: the queried 3D coordinate $\mathbf{x}$ is first scaled to match the native resolution of $l$, and the neighboring $2^3$ vertices from the fixed resolution 3D grid are identified. Each vertex of interest is mapped to an entry in the $l^{th}$ level feature array and the final feature value corresponding to $\mathbf{x}$ is obtained through tri-linear interpolation. This feature value is then passed into an implicit function represented by an MLP, mapping from feature space to density and RGB values.

\paragraph{Forgetting in explicit features.}Consider the case where $T$ matches the total number of vertices at each grid resolution, such that a 1:1 mapping exists between grid vertices and feature embeddings. In the continual setting, features are only updated when the corresponding voxel is visible in the training views of the current task, whereas other features remain constant, unaffected by catastrophic forgetting. This is in stark contrast to the global updates observed in fully implicit representations such as in NeRF~\cite{Mildenhall2021NeRF}. In NeRF, each network parameter influences radiance and density values at every point in 3D space, and training on new data points overwrites information learned in the entire scene, even for regions not visible in the current training views.
\paragraph{Hashed feature tables.}In an effort to lower memory usage at higher grid resolutions, Instant-NGP proposes a hashed encoding scheme. At finer levels, a hash function $h: \mathcal{Z}^d \mapsto \mathcal{Z}_T$ is used to index into the feature array, effectively acting as a hash table. Following prior work~\cite{Mller2022InstantNG}, we use a spatial hash function of the form
\begin{equation}
    h(\mathbf{x}) = \left( \bigoplus_{i=1}^d x_i \pi_i \right) \;\text{mod}\;T,
\end{equation}
where $\bigoplus$ represents the bit-wise XOR operator and $\pi_i$ are unique, large primes. 

In contrast to dense feature grids, hashed feature tables suffer from catastrophic forgetting in the feature space due to hash collisions. Consider a single task $\mathcal{I}_i$. A vertex $v_1$ visible in $\mathcal{I}_i$ may share the same hash table entry as another vertex $v_2$ that is not visible in $\mathcal{I}_i$. During training, the training objective will only optimize the shared hash table entry for the current task $\mathcal{I}_i$, learning the correct feature value for $v_1$, while forgetting any information learnt for $v_2$. The effects of forgetting are dependent on the frequency of hash collisions between grid vertices, which increases as the hash table size $T$ decreases.


\subsection{Memory replay through NeRF distillation}\label{distillation}
Catastrophic forgetting results from a misalignment between the current and cumulative training objectives. Replay-based approaches~\cite{Klein2007ParallelTA,Rebuffi2016iCaRLIC,Rolnick2018ExperienceRF,Shin2017ContinualLW,LopezPaz2017GradientEM,Chaudhry2018EfficientLL} combat network forgetting by storing information from previous tasks either explicitly or implicitly through a generative model. Consider a NeRF with explicit feature embeddings trained on a set of tasks $\{\mathcal{I}_1, \dots, \mathcal{I}_i\}$, with feature and MLP parameters characterized by $(\Phi_i, \Theta_i)$. We can then treat $(\Phi_i, \Theta_i)$ as a generator for 2D image data found in tasks $\{\mathcal{I}_1, \dots, \mathcal{I}_i\}$. Let $\hat{\mathbf{R}}_{i}$ be the union of all ground truth rays in the first $i$ tasks. The ground truth RGB value corresponding to a ray $\mathbf{r} \in \hat{\mathbf{R}}_{i}$ can then be approximated by $\hat{\mathbf{C}}(\mathbf{r};\Phi_i, \Theta_i)$ following Eq.~\ref{eq:vol_render}.

We approach continual learning in a self-distillation manner~\cite{Hinton2015DistillingTK}. When training on the subsequent task $\mathcal{I}_{i+1}$, we no longer have access to ground truth image data from previous tasks. However, as explored in prior work~\cite{Chung2022MEILNeRFMI}, by saving network parameters $(\Phi_i, \Theta_i)$ we effectively have access to pseudo ground truth values for all rays in $\hat{\mathbf{R}}_{i}$. We can then modify our training objective to minimize photometric loss for all rays in tasks $\{\mathcal{I}_1, \dots, \mathcal{I}_{i+1}\}$, rather than just $\mathcal{I}_{i+1}$. The modified training objective is then given by
\begin{align}
    \mathcal{L}(\Phi,\Theta)_{i+1} & = \sum_{\mathbf{r} \in \mathcal{I}_{i+1}} || \hat{\mathbf{C}}(\mathbf{r};\Phi, \Theta) - \mathbf{C}(\mathbf{r})||^2 \nonumber \\
    & + \sum_{\mathbf{r} \notin \mathcal{I}_{i+1}} || \hat{\mathbf{C}}(\mathbf{r};\Phi, \Theta) - \hat{\mathbf{C}}(\mathbf{r};\Phi_{i}, \Theta_{i})||^2.
    \label{loss}
\end{align}
During each task, we still sample rays uniformly over all previous and current tasks. However, for previous tasks where ground truth RGB values are no longer available, we instead query the frozen network to obtain a pseudo ground truth value. Figure~\ref{fig:method} shows a visualization of the replay-based distillation method.

\begin{figure*}[t]
    \centering
    \includegraphics[width=\linewidth]{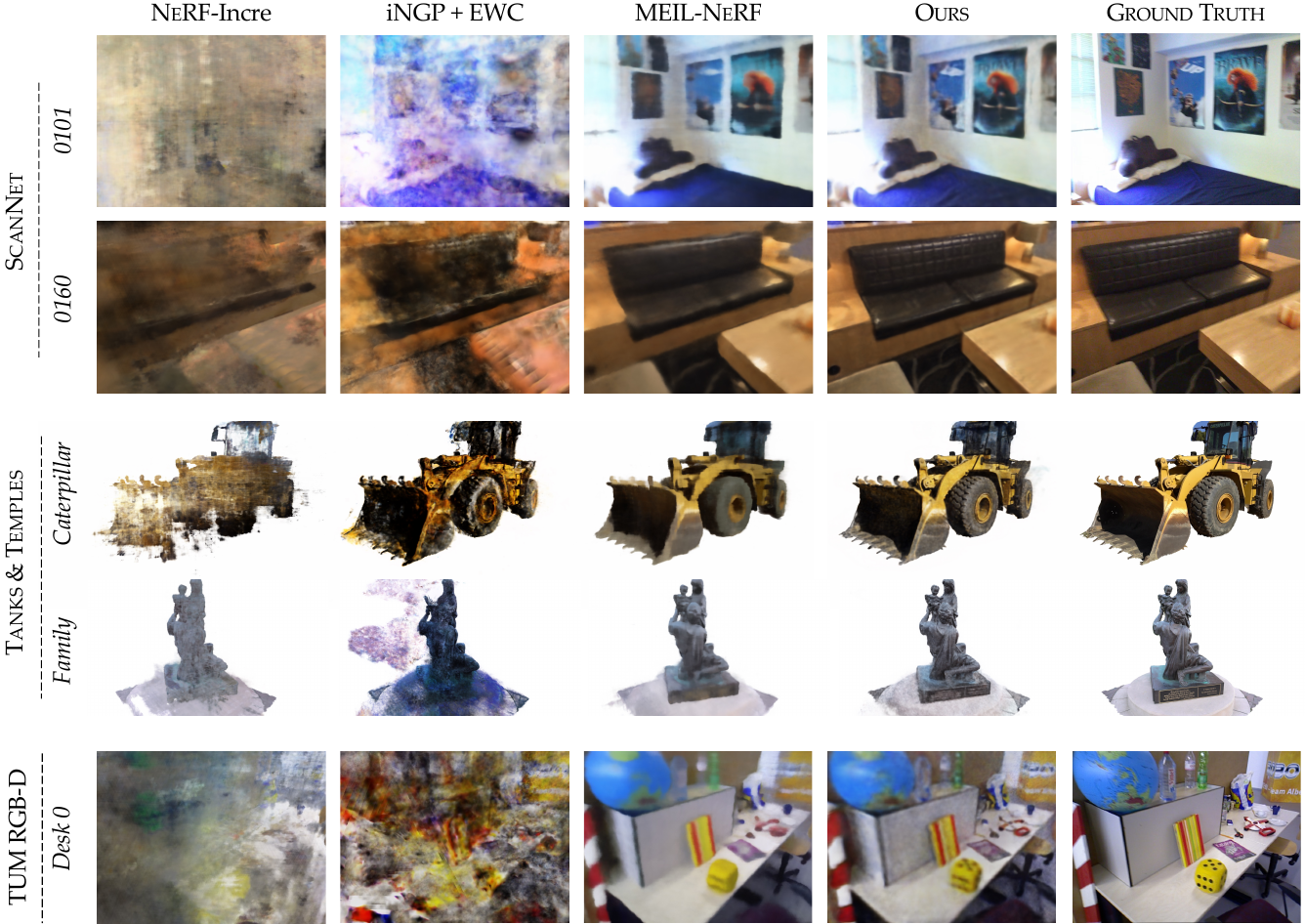}
    \caption{\textbf{Qualitative results: unconstrained setting.} We show reconstructed views from a previously supervised (forgotten) task across different methods. Our method consistently outperforms all other baselines in visual quality. NeRF trained in a continual setting suffers from catastrophic forgetting, as illustrated by poor early-task reconstruction results. Parameter regularization through EWC aids in alleviating forgetting effects, however, reconstruction results  still suffer from severe visual artefacts. MEIL-NeRF adopts a similar replay approach as our method, using a frozen copy of the scene representation as guidance when training future tasks. However, the fully implicit representation in MEIL-NeRF forgets high-frequency detail from earlier tasks. In contrast, our method is able to retain high-frequency details for earlier tasks through the use of explicit features.}\label{fig:unconstrained}
    \vspace{-1em}
\end{figure*}

\section{Experiments}\label{experiments}
To highlight the effectiveness of our method in overcoming catastrophic forgetting, we compare our method against existing continual learning methods~\cite{Kirkpatrick2016OvercomingCF, Chung2022MEILNeRFMI}. We describe baseline methods in Section~\ref{baselines}, datasets used in Section~\ref{datasets} and experimental settings in Section~\ref{settings}.

\subsection{Baselines}\label{baselines}

\paragraph{NeRF and iNGP.}We train NeRFs under the continual setting using frequency and multi-resolution hash encodings, referring to these baselines as \textit{NeRF-Incre} and \textit{iNGP-Incre} respectively. For our hash encoding experiments, we used a feature grid of $L = 16$ levels, a hash table size of $T = 2^{17}$, a feature dimension of $F=2$ and grid resolutions ranging from 16 to 512. We also scale the original NeRF representation~\cite{Mildenhall2021NeRF} to have 8 fully connected layers with 512 channels each, matching the total number of trainable parameters as the hash encoding models.

\paragraph{Elastic Weight Consolidation.}Elastic Weight Consolidation (EWC)~\cite{Kirkpatrick2016OvercomingCF} is a form of feature regularization method for alleviating catastrophic forgetting. Let $\Phi_A$ be the set of hashed feature embeddings learned on task $\mathcal{I}_A$. Consider a subsequent task $\mathcal{I}_B$. EWC modifies the training objective to the following:
\begin{equation}
    \mathcal{L}(\Phi) = \mathcal{L}_B(\Phi) + \frac{\lambda}{2} F (\Phi - \Phi_A)^2.
\end{equation}
$\mathcal{L}_B$ represents the training objective on task $\mathcal{I}_B$ and $F$ is an estimation of the diagonal of the Fischer information matrix given by the squared gradients of parameters $\Phi_A$ with respect to the training objective $\mathcal{L}_A$. Intuitively, $\Phi_A$ is recorded as a set of reference parameters. Deviation from these reference parameters are penalized, weighted on their importance relative to the training objective. We implement EWC on top of an iNGP backbone as a baseline method by fixing the trained network parameters after each training task as the reference parameters.

\paragraph{MEIL-NeRF.} Recently, MEIL-NeRF~\cite{Chung2022MEILNeRFMI} also proposed the use of memory replay through network distillation for alleviating catastrophic forgetting effects in NeRFs. However, MEIL-NeRF uses the original fully implicit NeRF representation as a backbone, which limits reconstruction quality and convergence speed. We include continual learning results following the general implementation of MEIL-NeRF. While MEIL-NeRF uses an additional ray generator network for sampling previous rays from previous tasks, this additional step leads to significant degradation in reconstruction results while providing minimal memory savings; we therefore omit this step and sample ground truth rays instead. MEIL-NeRF also explores using Charbonnier penalty function, we consider changes to the penalty function a tangential area of exploration, and choose to train both our method and MEIL-NeRF using the loss function detailed in Equation~\ref{loss}.

\subsection{Datasets}\label{datasets}
We compare methods on the task of continual scene fitting using the Tanks \& Temples~\cite{Knapitsch2017TanksAT}, ScanNet~\cite{Dai2017ScanNetR3} and TUM RGB-D datasets~\cite{Sturm2012ABF}. Data for each scene is represented by a trajectory of ground truth camera poses and corresponding RGB images, with each trajectory containing 100--300 images depending on scene. We emulate the setting of continual learning by partitioning each trajectory into 10 temporally sequential tasks.

\rowcolors{2}{light-gray}{}
\begin{table*}[ht!]
  \centering
  \begin{tabular}{m{10em}m{4em}m{3.5em}m{3.5em}m{3.5em}m{4.5em}m{3.5em}m{3.5em}m{3.5em}}
  \toprule
    & \multicolumn{3}{c}{\textit{ScanNet}} & \multicolumn{3}{c}{\textit{Tanks \& Temples}} & \multicolumn{2}{c}{\textit{TUM RGB-D }}\\
     \;\;\textit{Method} & \textit{0101}  & \textit{0146} & \textit{0160} & \textit{Truck} & \textit{Caterpillar} & \textit{Family} & \textit{TUM 1} & \textit{TUM 2}\\
    \midrule
    \;\;Ours (1 s) & 19.61 & 22.18 & 23.84  & 19.19 & 17.24 & 23.24 & 15.05 & 16.65 \\
    \;\;Ours (5 s)  & 24.10 \bronze& 26.13 \bronze& 28.37 \bronze& 21.93 \bronze& 20.59 \bronze& 26.29 \bronze& 19.35 \bronze& 19.02 \bronze\\
    \;\;Ours (30 s) & 25.54 \gold& 27.84 \gold& 30.45 \gold& 23.97 \gold& 22.62 \gold& 29.21 \gold& 21.09 \gold& 20.42 \gold\\
    \midrule
    \;\;MEIL--NeRF (30 s) & 18.85 & 21.41 & 22.72 & 18.11 & 16.93 & 21.78  & 16.05 & 15.96 \\
    \;\;MEIL--NeRF (1 min) & 20.65 & 22.76 & 24.39 & 19.38 & 18.41 & 23.19& 17.44 & 16.19\\
    \;\;MEIL-NeRF (10 min) & 24.32 \silver& 26.82 \silver& 28.93 \silver& 22.74 \silver& 20.89 \silver& 26.57 \silver& 20.79 \silver& 19.80 \silver\\
    \bottomrule
  \end{tabular}
    \vspace{0.2em}
    \caption{\textbf{Quantitative results: time constrained.} We show reconstruction PSNR for our method and MEIL-NeRF trained on a fixed time limit per task.
    Our method converges to better results at a much faster rate. Our method trained for only 5s per task outperforms MEIL-NeRF trained for 1 min per task and is competitive with MEIL-NeRF trained for 10 min per task. Given its rapid convergence, our method uniquely enables real-time continual scene reconstruction.}\label{tab:speed_eval}
\end{table*}

\begin{figure*}[t]
    \centering
    \includegraphics[width=\linewidth]{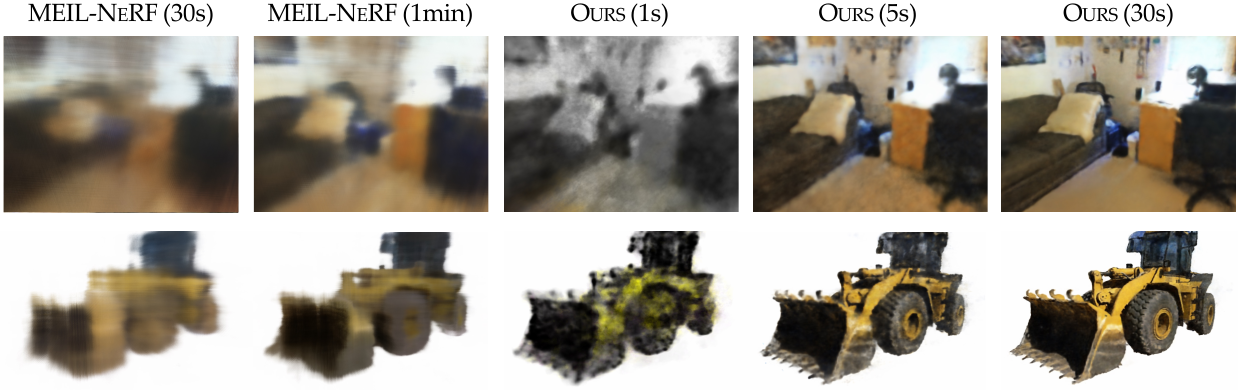}
    \caption{\textbf{Qualitative results: time constrained.} We show reconstructed views from an earlier supervised (forgotten) task for our method and MEIL-NeRF trained for fixed times per task. Our method consistently outperforms MEIL-NeRF given equal time budget. With only 5s per task, our method already reconstructs the scene with reasonable fidelity, illustrating that our method is well-suited for real-time continual scene fitting.}\label{fig:speed_qual}
\end{figure*}

\subsection{Experimental settings}\label{settings}
We evaluate our method in two separate settings: an unconstrained setting where each method is trained on every task until convergence, and a constrained setting where each task is trained on a fixed time budget. The unconstrained setting aims at testing the upper-bound performance of each method, while the constrained setting mimics a real-time continual scene reconstruction setting. Each model is trained on a single RTX 6000 GPU, with a ray batch size of 1024. For the unconstrained settings, we trained methods using a hash encoding for 1 minute per task and methods built on fully implicit NeRFs for 10 minutes per task. 

\subsection{Results}\label{results}
\paragraph{Unconstrained setting.}We show quantitative results of each method for the unconstrained setting in Table~\ref{tab:main_results}. Methods are evaluated using peak signal-to-noise ratio (PSNR), averaged over all images in the test trajectory. We also provide quantitative results of the fully implicit NeRF and hash-encoding representations trained in a non-continual setting. These results serve as an upper bound for their continual learning counterparts. Quantitatively, our method consistently outperforms all baselines in reconstruction quality. While performance of MEIL-NeRF comes close to our method for certain scenes, our method takes significantly less time to train due to the convergence properties of the hash encoding representation. Results from our method also come very close to the theoretical upper bound set by the results obtained from non-continual training, further illustrating the effectiveness of our method.

Figure~\ref{fig:unconstrained} shows qualitative results from the unconstrained setting. Naively training NeRF under the continual setting leads to catastrophic forgetting, as earlier views contain heavy artefacts. Parameter regularization through EWC helps alleviate forgetting for certain scenes, however, reconstruction quality is still limited. MEIL-NeRF produces visually pleasing results, but reconstruction of earlier views lack high-frequency details. In contrast, our method is able to retain these high frequency details, as the underlying multi-resolution hash encoding stores high-frequency features explicitly, allowing high frequency details to be retained during training.
    
\paragraph{Time-constrained setting.} We evaluate our method against MEIL-NeRF in the time-constrained setting. We trained both methods on each task for a fixed period of time and show reconstruction PSNR averaged over all views along the test trajectory in Table~\ref{tab:speed_eval}.  Our method trained on 30 seconds per task out performs MEIL-NeRF, even when trained for 10 minutes per task. More importantly, our method trained for just 5 seconds produces results comparable to MEIL-NeRF at convergence. Qualitative results in Figure~\ref{fig:speed_qual} show that our method provides reasonable scene reconstruction quality at much shorter training times, illustrating that our method is uniquely suited for the task of real-time continual scene fitting.

\subsection{Ablation study}

\begin{figure}[t]
    \centering
    \includegraphics[width=\linewidth]{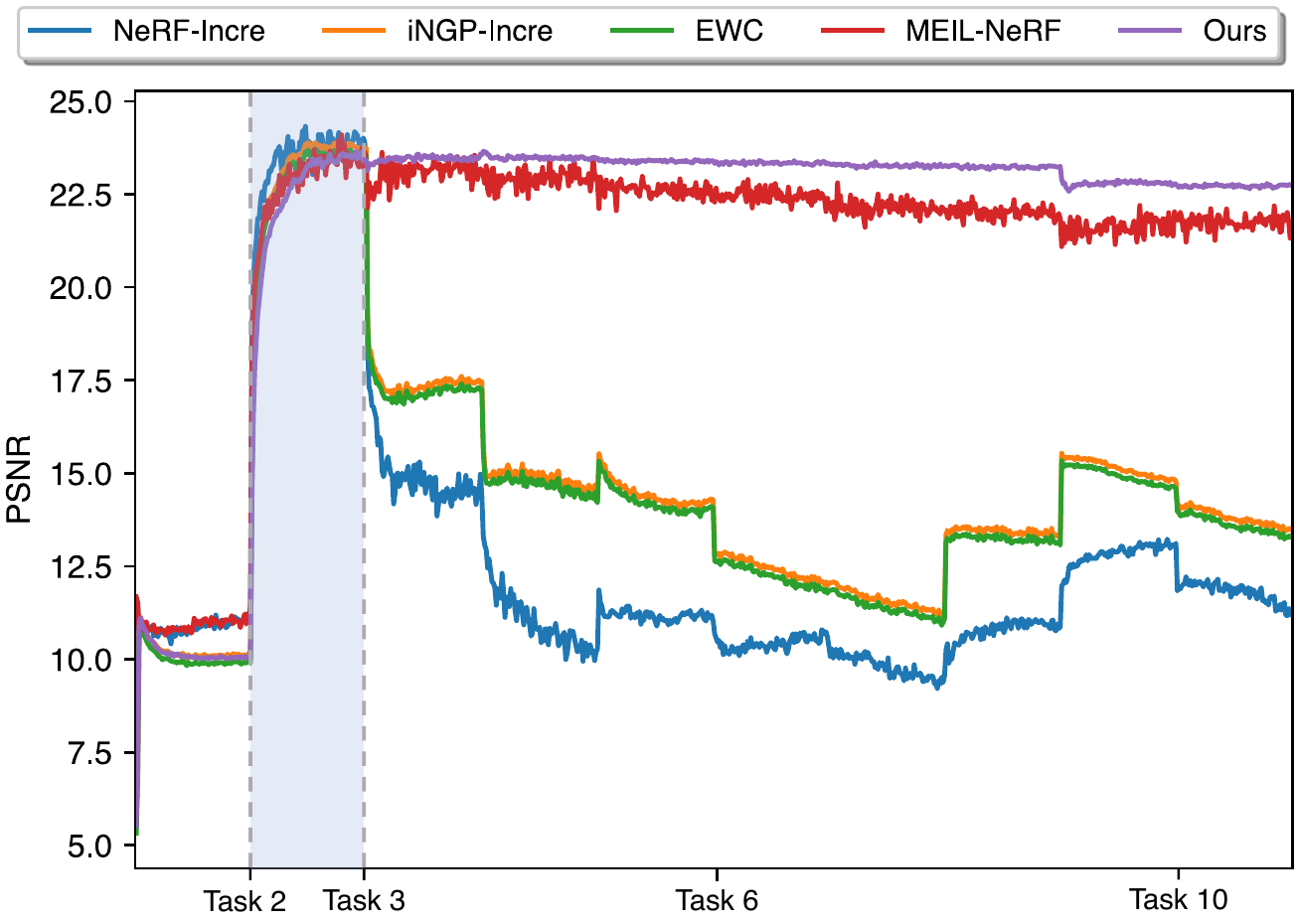}
    \caption{\textbf{Reconstruction quality of early-supervised tasks.} Reconstruction PSNR of task 2 over the course of training. Our method successfully alleviates degradation effects from catastrophic forgetting, and consistently outperforms all other baselines.}\label{fig:degradation}
    \vspace{-1.5em}
\end{figure}

\paragraph{Degradation of early tasks.} We evaluate reconstruction PSNR of the second task (of ten total) over the course of training using different methods in Figure~\ref{fig:degradation}. Conventional methods naively trained in the continual setting (\textit{NeRF-Incre \& iNGP-Incre}) experience severe degredation due to catasatrophic forgetting. MEIL-NeRF succesfully alleviates forgetting effects through self-distillation, however, forgetting effects are still observed after training for many tasks. In contrast, our method is able to maintain high PSNR for previous tasks even after training for many tasks.

\subsection{Applications: autonomous vehicle data}
\begin{figure}[t]
    \centering
    \includegraphics[width=\linewidth]{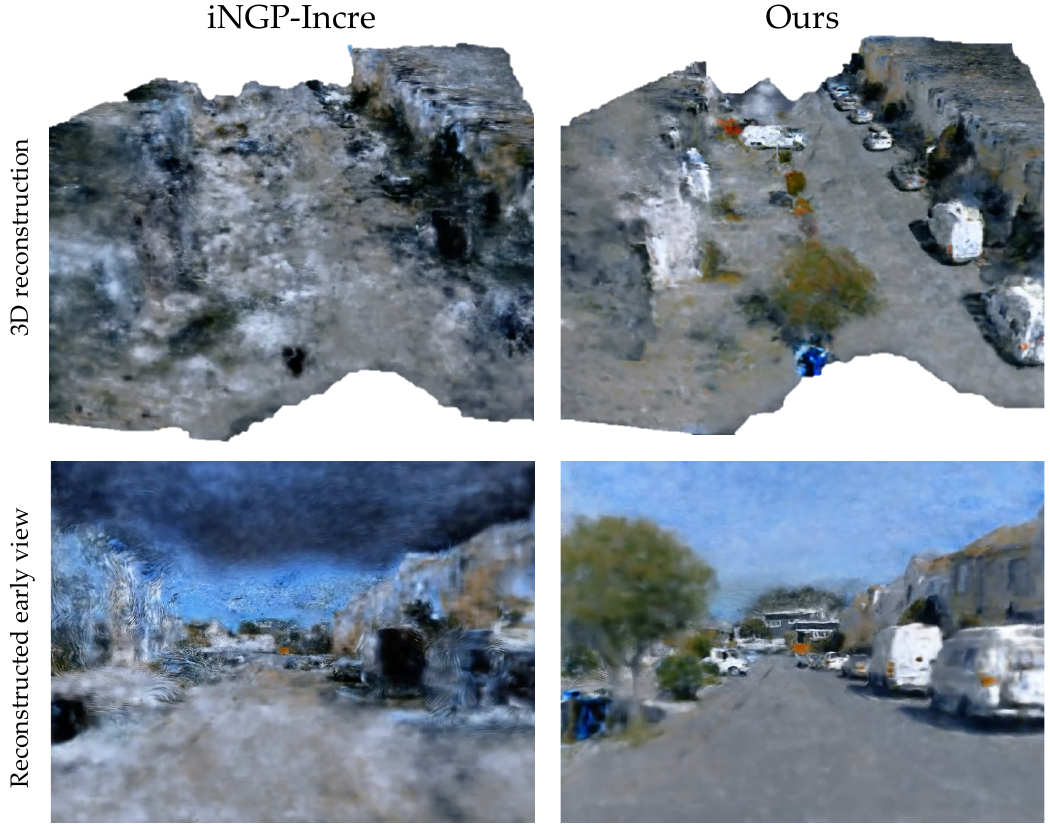}
    \caption{\textbf{Qualitative results on Waymo open dataset.} Our method recovers earlier training views at higher quality than training NeRFs naively in a continual learning setting.}\label{fig:waymo}
    \vspace{-1.5em}
\end{figure}

Our method is well-suited for scenarios such as autonomous vehicle captures and drone footage, where data is sequentially acquired and an updated 3D scene representation should be immediately available. To illustrate this, we train our method on data obtained from the Waymo open dataset~\cite{Sun2019ScalabilityIP}. A single trajectory in the Waymo dataset consists of a video stream from 5 calibrated cameras mounted at the top of the vehicle. Similar to the experimental settings described in Section~\ref{experiments}, we split each trajectory into 10 temporally sequential tasks. We show qualitative results using our method and iNGP-Incre in Figure~\ref{fig:waymo}, training each task for 30 seconds for a total of 5 minutes. Our method recovers meaningful geometry and reconstructs earlier views with much higher quality.


\section{Discussion}
\paragraph{Limitations and future work.} Our method relies on ground truth camera poses to perform scene fitting. Although prior works have explored simultaneous optimization of camera poses and scene parameters for NeRFs, they either rely on good initializations~\cite{Lin2021BARFBN} or specific constraints on the distribution of camera poses~\cite{Levy2023MELONNW}. Simultaneous estimation of camera poses in the setting of continual learning for NeRFs has also yet to be explored. It may be fruitful to explore this direction further in relation to methods for SLAM~\cite{Bailey2006SimultaneousLA}.

We chose to use multi-resolution hash encodings~\cite{Mller2022InstantNG} to leverage its fast convergence properties and explicitly defined features to combat forgetting. Alternate representations, such as triplanes~\cite{Chan2021EfficientG3} and TensoRF~\cite{Chen2022TensoRFTR} can also be explored as potential substitutes, potentially further increasing robustness to catastrophic forgetting through more structured encodings.

Our method uses a frozen version of the scene representation network trained on previous tasks as a pseudo ground truth oracle. Querying the network for pseudo ground truth values requires volume rendering through the scene, adding computational overhead to the training process. A potential direction of exploration is to find other forms of compression, such as 2D coordinate networks~\cite{Sitzmann2020ImplicitNR}, to act as the pseudo ground truth oracle. Additionally, if any single oracle network is not of sufficient quality, this will continue to affect downstream training on subsequent tasks.

\paragraph{Conclusion.}
In this work, we aim to extend the practical viability of NeRFs, specifically in the continual setting, where training data is sequentially captured and a 3D representation needs to be immediately available. By combining multi-resolution hash encodings and replay methods through network distillation, our approach alleviates the effects of catastrophic forgetting observed in the continual learning of NeRFs. While previous approaches struggle with quality and speed, our method is able to produce visually compelling reconstruction of earlier tasks while being an order of magnitude faster than existing methods.

\section{Acknowledgements}
We thank Geoff Burns, Abhinav Modi, Jing Cui, and Hamid Izadi Nia for invaluable discussions and feedback. This project was in part supported by Rivian, the Samsung GRO program and a PECASE from the ARO. Ryan Po is supported by the Stanford Graduate Fellowship.
{\small
\bibliographystyle{ieee_fullname}
\bibliography{egbib}
}
\newpage
\onecolumn
\appendix
\noindent{\Large \textbf{{Appendix}}}

\section{Additional Results}

\subsection{Degradation of early task}
We show additional qualitative results on the degradation of earlier views under the continual learning setting for each method in Supp. Fig. \ref{fig:task2-deg}. Naive methods such as NeRF-Incre and EWC suffer from catastrophic forgetting. Although the early view is reconstructed well when the current view is part of the training views of the current task, reconstruction quality quickly degrades as the training task moves to other regions of the 3D scene. Our method and MEIL-NeRF \cite{Chung2022MEILNeRFMI} maintains relatively high quality throughout all stages of training.
\begin{figure}[h!]
    \centering
    \includegraphics[width=\linewidth]{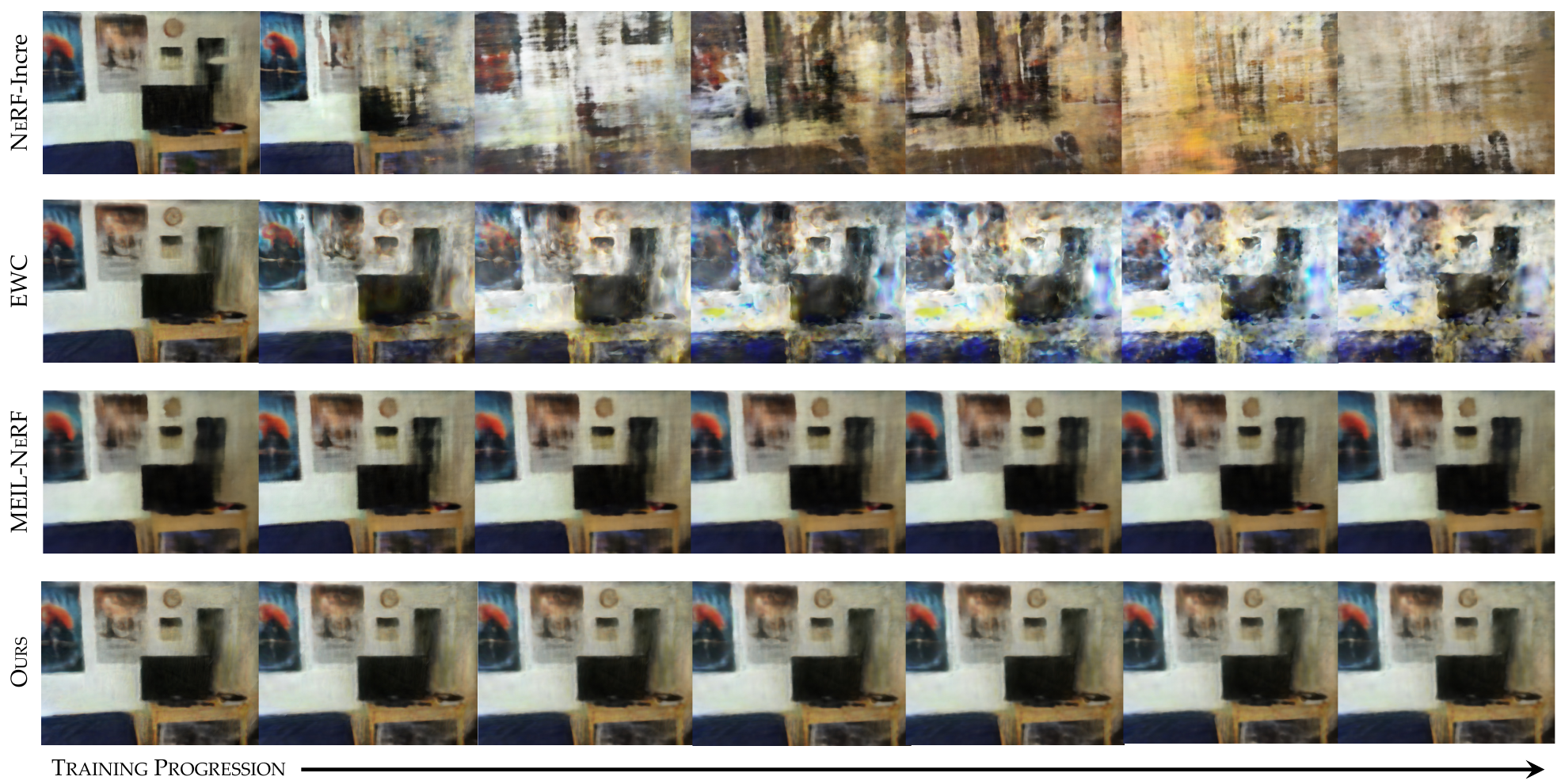}
    \caption{Reconstructed view of earlier task during different stages of training. Quantitative results shown in Figure 6 are obtained from evaluating PSNR of the right-most column.}
    \label{fig:task2-deg}
    \vspace{-1.5em}
\end{figure}

\newpage

\subsection{Ablation on \# of tasks}
We perform an ablation over the number of tasks used for the continual learning setting, showing results in Table \ref{tab:num-tasks}. As the number of tasks increases, the continual learning setting becomes more challenging, causing performance of our method to drop minimally as number of tasks increases. Naive method of NeRF-Incre performs poorly in the continual learning setting even for lower number of tasks.
\begin{table}[h!]
  \centering
  \small
  \begin{tabular}{ccccc}
    \toprule
     & \multicolumn{4}{c}{\textit{\# of Tasks}} \\
     
     Method & 2 & 5 & 10 & 20\\
    \midrule
    NeRF-Incre &
12.30& 16.40& 13.70& 13.43 \\
    Ours & 24.29& 24.27 & 24.13 & 23.87\\
    \bottomrule
  \end{tabular}
  \caption{Ablation on \# of tasks in continual learning setting. Results obtained from 0101 scene of the ScanNet dataset \cite{Dai2017ScanNetR3}.}
  \label{tab:num-tasks}
\end{table}

\subsection{Reconstruction quality for individual tasks}
Table \ref{tab:all-tasks} shows reconstruction PSNR for scene ScanNet-0101 for every single task after training has concluded. As expected, non-replay baselines (NeRF-Incre, iNGP-Incre, EWC) suffer from catastrophic forgetting, where earlier tasks are reconstructed poorly, but having high reconstruction PSNR for the final task. Replay based methods such as MEIL-NeRF and ours maintain high PSNR across all tasks, with our method achieving better performance across all tasks.
\begin{table}[h!]
  \centering
  \small
  \begin{tabular}{ccccccccccc}
    \toprule
    Method & Task 1 & Task 2 & Task 3 & Task 4 & Task 5 & Task 6 & Task 7 & Task 8 & Task 9 & Task 10 \\
    \midrule
    NeRF-Incre &
13.92& 13.17& 11.01& 10.94& 12.04& 13.84& 15.96& 18.73& 17.43& 25.01 \\
iNGP-Incre &
14.86& 15.02& 12.89& 15.15& 16.50& 18.21& 18.40& 21.07& 18.68& 25.37\\
EWC &
13.79& 14.07& 12.66& 14.66& 16.50& 18.09& 18.46& 20.64& 17.98& 25.34\\
MEIL-NeRF &
23.84& 24.29& 22.95& 22.85& 24.10& 24.72& 25.47& 26.31& 25.57& 24.68\\
    Ours & 24.19 & 26.33 &24.74&24.32&26.12&26.08&26.78&27.15&25.96&24.92 \\
    \bottomrule
  \end{tabular}
  \caption{Reconstruction PSNR of individual tasks after completion of training over all tasks in the continual learning setting. }
  \label{tab:all-tasks}
\end{table}

\section{Code and Implementation Details.}  
We implement our method as a modular component of the Nerfstudio pipeline \cite{Tancik2023NerfstudioAM}. Our code includes custom data parsers for processing datasets to be suited for the continual learning setting described in the paper. All code and custom data formats will be made available upon acceptance of the paper.

\begin{figure}[h!]
    \centering
    \includegraphics[width=0.7\linewidth]{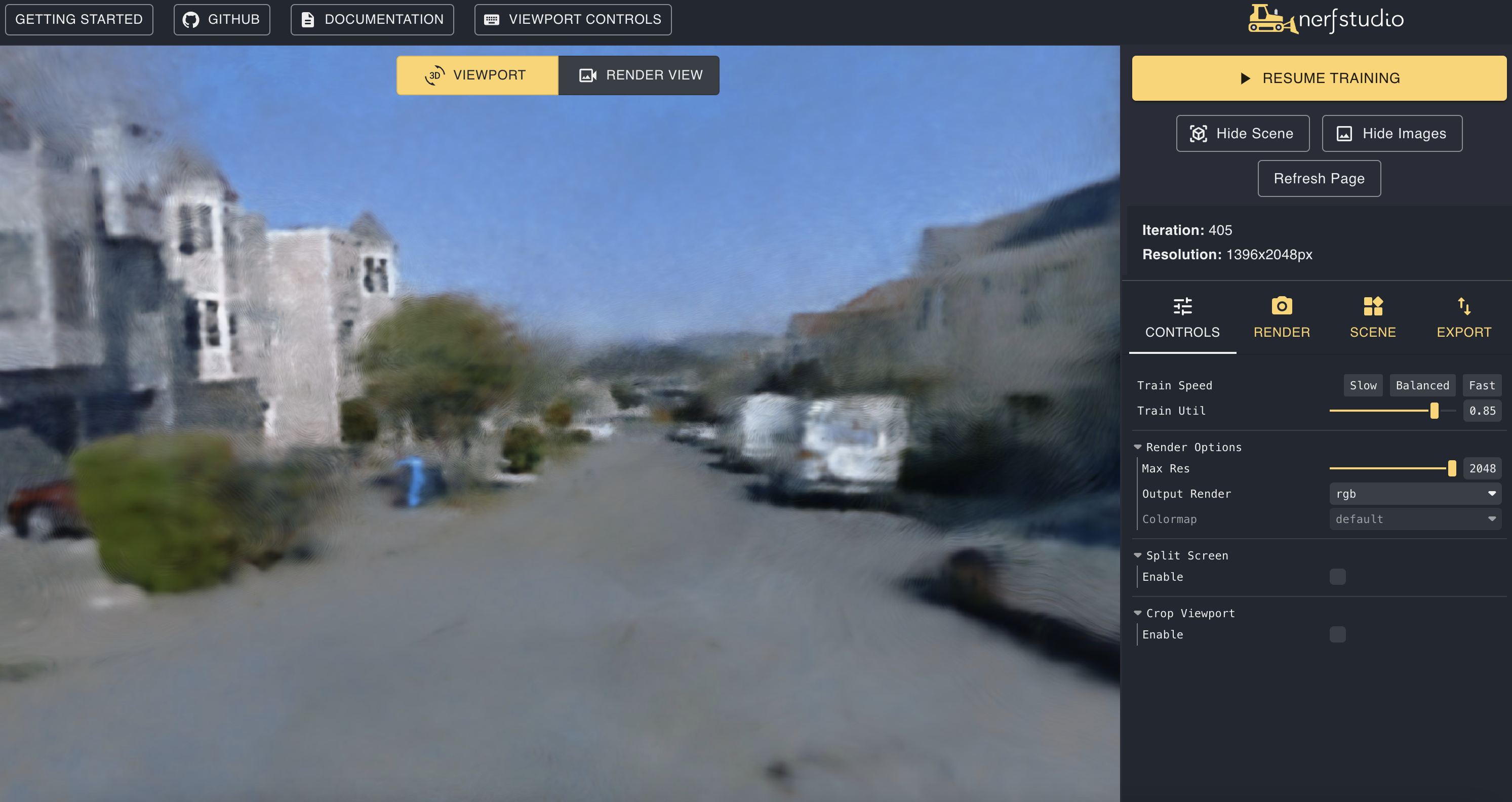}
    \caption{Nerfstudio interface for continual learning of NeRFs. Current view shows training for an initial task from the Waymo open dataset \cite{Sun2019ScalabilityIP}.}
    \label{fig:task2-deg}
    \vspace{-1.5em}
\end{figure}

\end{document}